\documentclass[graybox]{svmult}

\usepackage{mathptmx}       
\usepackage{helvet}         
\usepackage{courier}        
\usepackage{type1cm}        
\usepackage{makeidx}         
\usepackage{graphicx}        
\usepackage{multicol}        
\usepackage[bottom]{footmisc}
\usepackage{booktabs}
\usepackage{amsmath}
\usepackage{subcaption}

\begin{document}
\title*{PhilHumans: Benchmarking Machine Learning for Personal Health}

\author{Vadim Liventsev$^5$, Vivek Kumar$^6$, Allmin Pradhap Singh Susaiyah$^5$, Zixiu Wu$^6$ , Ivan Rodin$^4$, Asfand Yaar$^4$, Simone Balloccu$^3$, Marharyta Beraziuk$^6$, Sebastiano Battiato$^4$, Giovanni Maria Farinella$^4$, Aki H\"arm\"a$^7$, Rim Helaoui$^7$,  Milan Petkovic$^5$, Diego Reforgiato Recupero$^6$, Ehud Reiter$^3$, Daniele Riboni$^6$, and Raymond Sterling$^1$}


\institute{$^1$R2M Solution, Spain,
$^2$Joint Research Centre, Italy,  
$^3$University of Aberdeen, Scotland
$^4$University of Catania, Italy, 
$^5$TU Eindhoven, the Netherlands (Corresponding author's \email{v.liventsev@tue.nl}), 
$^6$University of Cagliari, Italy,
$^7$Philips Research, Eindhoven, the Netherlands.
The affiliations listed reflect those at the time the work was completed and may not be current.
}

%
%

\maketitle

\abstract{The use of machine learning in Healthcare has the potential to improve patient outcomes as well as broaden the reach and affordability of Healthcare. The history of other application areas indicates that strong benchmarks are essential for the development of intelligent systems. We present Personal Health Interfaces Leveraging HUman-MAchine Natural interactions (PhilHumans), a holistic suite of benchmarks for machine learning across different Healthcare settings - talk therapy, diet coaching, emergency care, intensive care, obstetric sonography - as well as different learning settings, such as action anticipation, timeseries modeling, insight mining, language modeling, computer vision, reinforcement learning and program synthesis}

\section{Introduction}
\label{sec:intro}

Understaffing has been consistently identified as the major challenge facing Healthcare today \cite{ashleyy.metcalfHospitalUnitUnderstaffing2016,SurveyShowsHidden1993,UnderstaffingSignificantIssue2012,campbell_universal_2013, hudsonUnderstaffing2015, mercerMessageEditorinChief2008, r.stanleyUnderstaffedOverwhelmed2010, munnUnderstaffingWardsCompromising2017, thelancetHealthcareSystemStaffing2018}. Automation tools that make use of Machine Learning (also known as Healthcare 4.0 \cite{tortorellaHealthcareTrendsChallenges2020}) have been consistently identified as crucial for reducing the workload of Healthcare professionals and improving the quality of care \cite{agrawalMachineLearningHealthcare2020, deviDesignImplementationAdvanced2022, g.kumarSurveyMachineLearning2016, ganguliMachineLearningPursuit2020, maityMachineLearningImproved2017, mitraMachineLearningHealthcare2021, pianykhImprovingHealthcareOperations2020, xhaferraRoleMachineLearning2022}. In turn, the shortage of standard benchmarks has been consistently identified as a central roadblock for machine learning in Healthcare
\cite{Crown2015Potential, David2020Evaluating, guSupervisedLearningPervasive2023, harutyunyanMultitaskLearningBenchmarking2019, Kathrin2022Benchmark, liventsevEffectivePatientSimulators2021, mcdermottReproducibilityMachineLearning2021, purushothamBenchmarkingDeepLearning2018, S2017Benchmark}.

Whether it's ImageNet \cite{dengImagenetLargescaleHierarchical2009} in Computer Vision or GLUE \cite{wangGLUEMultitaskBenchmark2018} in natural language processing, benchmarks are a core research tool in mature applications of machine learning, enabling quantitative analysis of learning methodologies to guide and orient their development.
Machine learning for Healthcare, an emergent field with unique challenges in availability of research datasets \cite{Anshik2021Handling, Gilbert2015market, Pahwa2021Big, Yazhini2019State} lacks an accepted benchmarking standard: recent literature reviews \cite{palMachineLearningHealthcare2023,tortorellaHealthcareTrendsChallenges2020} of the field cover a variety of studies that each use their own (often non-public) benchmark.

To address this, we propose a suite of tasks that each include a dataset and an evaluation procedure and cover a wide variety of both healthcare settings (therapy and coaching, emergency care, intensive care, sonography) and machine learning challenges (conversational agents, computer vision, time series prediction, reinforcement learning). We also provide a preliminary evaluation of widely used machine learning approaches on these tasks.

\section{Benchmarks}

\subsection{Tabular perspective}

Our first two benchmarks evaluate methods for machine learning on low-dimensional tabular data.
This perspective restricts the types of data made available to learning algorithms, excluding for example, radiological data. 
At the same time, it is anything but contrived: many real-world tasks, such as risk assessment for negative outcomes in intensive care based on the patient's vital signs, are within the tabular domain \cite{reynaEarlyPredictionSepsis2020}.

\subsubsection{MIMIC-IV-Ext-SEQ}

Reinforcement Learning in Healthcare is typically concerned with narrow self-contained tasks such as sepsis prediction or anesthesia control.
However, previous research has demonstrated the potential of generalist models \cite{reedGeneralistAgent2022} (the prime example being Large Language Models \cite{brown2020:language}) to outperform task-specific approaches due to their capability for implicit transfer learning.
To enable training of foundation models for Healthcare as well as leverage the capabilities of state of the art Transformer architectures, we propose the paradigm of Healthcare as Sequence Modeling, in which interaction between the patient and the healthcare provider is represented as an event stream and tasks like diagnosis and treatment selection are modeled as prediction of future events in the stream. 
To explore this paradigm experimentally we develop MIMIC-IV-Ext-SEQ~\cite{liventsevIntensiveCareOne2024}, a soon-to-be-published sequence modeling benchmark derived by translating heterogeneous clinical records from MIMIC-IV~\cite{johnsonMIMICIVFreelyAccessible2023} dataset into a uniform event stream format.
Information from different sources in the MIMIC-IV clinical database is represented as an offline reinforcement learning trajectory of interactions between the doctor and the patient where every is a word from a 87899-word interaction vocabulary.

\subsubsection{Auto-ALS}

\begin{figure}
    \centering
    \includegraphics[width=\linewidth]{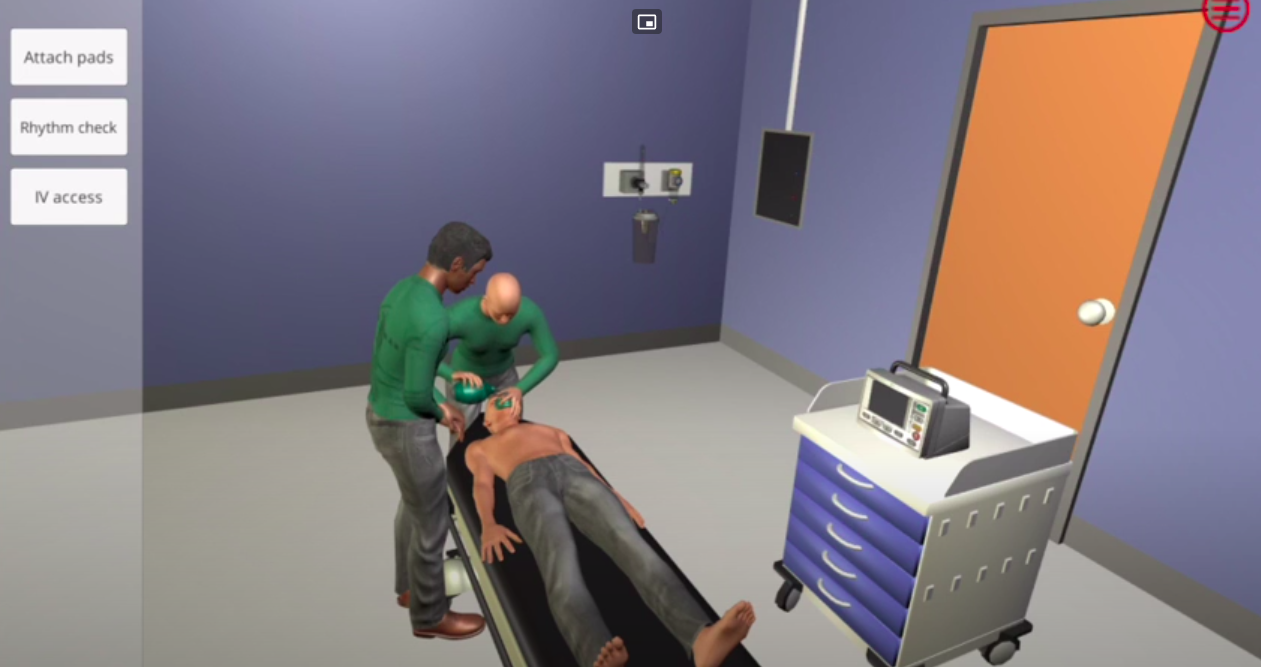}
    \caption{Virtu-ALS}
    \label{fig:virtu-als}
\end{figure}

Virtu-ALS is an emergency care simulator designed to train students and junior healthcare professionals in following the ABCDE Assessment Protocol \cite{thimInitialAssessmentTreatment2012} for emergency care, although its application as a reinforcement learning \emph{benchmark} was anticipated and accounted for by the authors \cite{briskAIEnhanceInteractive2018}.
For a benchmark task, we add an Application Programming Interface to \emph{Virtu-ALS} and propose \emph{Auto-ALS} \cite{liventsevEffectivePatientSimulators2021}: a Gymnasium\footnote{https://gymnasium.farama.org/} environment\footnote{https://github.com/vadim0x60/auto-als} for training automated emergency care agents with Reinforcement Learning.
This is achieved by attaching an event listener to Virtu-ALS that registers all observable events that can occur in the simulator in response to the user's actions.

At each step of the episode the agent receives an observation vector of size 33+7+7=47 that represents various events that happened in the environment prior to the moment when `observation` is received. It implements an event encoding strategy described in \cite{liventsevReinforcementLearningMessage2021}.

The first 33 elements indicate whether one of 33 types of events has occured and how long ago it happened. For i=1,2,...,33:

\begin{equation}
o_i = \exp(t_i - t)
\end{equation}

i.e. the inverse exponent of how much time has passed since this event has last occured that can be interpreted as \emph{relevance} of this event at the moment. If the event has never occurred,  

\begin{equation}
    t - t_i = \infty \Rightarrow \exp(t_i - t) = 0
\end{equation}

The 33 events are as follows: 

\texttt{
    ResponseVerbal,
    ResponseGroan,
    ResponseNone,
    AirwayClear,
    AirwayVomit,
    AirwayBlood,
    AirwayTongue,
    BreathingNone, \\
    BreathingSnoring,
    BreathingSeeSaw,
    BreathingEqualChestExpansion,
    BreathingBibasalCrepitations,
    BreathingWheeze, \\
    BreathingCoarseCrepitationsAtBase,
    BreathingPneumothoraxSymptoms,
    VentilationResistance,
    RadialPulsePalpable,
    RadialPulseNonPalpable,
    HeartSoundsMuffled,
    HeartSoundsNormal,
    AVPU\_A,
    AVPU\_U,
    AVPU\_V,
    PupilsPinpoint,
    PupilsNormal,
    ExposureRash, \\
    ExposurePeripherallyShutdown,
    ExposureStainedUnderwear,
    HeartRhythm0,
    HeartRhythm1,
    HeartRhythm2,
    HeartRhythm3,
    HeartRhythm4
}

The next 7 components use the same time encoding $o_i = \exp(t_i - t)$ for vital signs measurement, i.e. how recently the last measurement has occured:

\texttt{
    MeasuredHeartRate,
    MeasuredRespRate,
    MeasuredCapillaryGlucose,
    MeasuredTemperature,
    MeasuredMAP,
    MeasuredSats,
    MeasuredResps
}

The last 7 components contain the measurements themselves.

After receiving the observation vector, the agent has to output an action: an integer no less than 0 and no more than 48. The 49 actions are:

\texttt{
    DoNothing,
    CheckSignsOfLife,
    CheckRhythm,
    ExamineAirway,
    ExamineBreathing,
    ExamineCirculation,
    ExamineDisability,
    ExamineExposure,
    ExamineResponse,
    GiveAdenosine,
    GiveAdrenaline,
    GiveAmiodarone,
    GiveAtropine,
    GiveMidazolam,
    UseVenflonIVCatheter,
    GiveFluids,
    ViewMonitor,
    StartChestCompression,
    OpenAirwayDrawer,
    OpenBreathingDrawer,
    OpenCirculationDrawer,
    OpenDrugsDrawer,
    BagDuringCPR,
    ResumeCPR,
    UseMonitorPads,
    UseSatsProbe,
    UseAline,
    UseBloodPressureCuff,
    AttachDefibPads,
    UseBagValveMask, \\
    UseNonRebreatherMask,
    UseYankeurSucionCatheter,
    UseGuedelAirway,
    TakeBloodForArtherialBloodGas,
    TakeRoutineBloods, \\
    PerformAirwayManoeuvres,
    PerformHeadTiltChinLift,
    PerformJawThrust,
    TakeBloodPressure,
    TurnOnDefibrillator,
    DefibrillatorCharge,
    DefibrillatorCurrentUp,
    DefibrillatorCurrentDown,
    DefibrillatorPace,
    DefibrillatorPacePause,
    DefibrillatorRateUp,
    DefibrillatorRateDown,
    DefibrillatorSync,
    Finish
}

Note, in particular, the `Examine` actions. These actions, just like `DoNothing` are guaranteed to have no effect on the patient state. However, some events will not trigger unless the agent goes looking for them. To check the blood pressure, one needs to attach the blood pressure cuff to the patient and look at the monitor. Hence, the `MeasuredMAP` event will only trigger after you `BPCuffOn` and `ExamineMonitor`. Examination skills are crucial for patient resusciation \cite{thimInitialAssessmentTreatment2012} - the simulation would be woefully inadequate if the Examinations were just provided automatically.

The episode ends when the agent outputs selects the \texttt{Finish} action. 
The episode is considered successful if the patient has been stabilized (has a clear airway, oxygen saturation of at least 88\%, a respiratory rate of at least 8 breaths per minute and a mean arterial pressure of at least 60mmHg).

\subsection{Vision perspective}

\subsubsection{Human-Robot interaction}

\begin{figure*}[t]
  \centering
  \includegraphics[width=1.\textwidth]{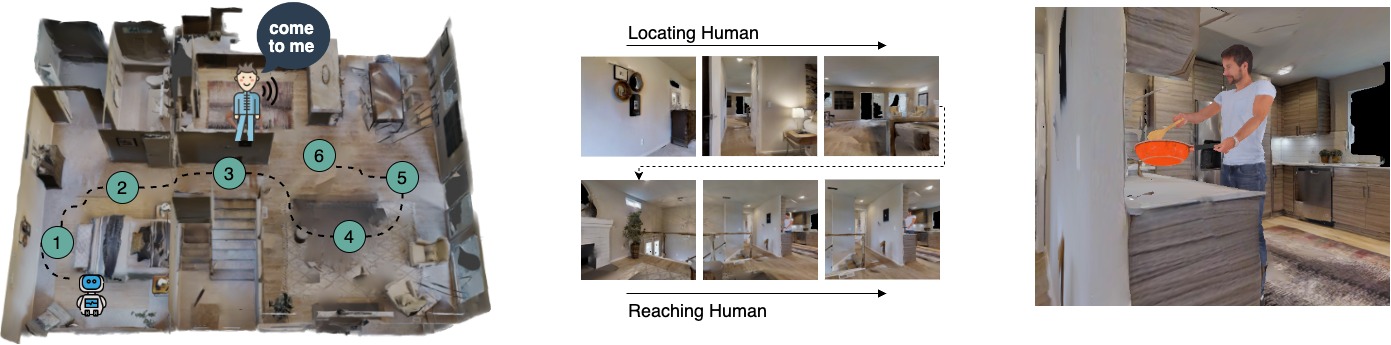}
  \caption{(left) Robot responding to a human call by exploring the environment with the help of multiple global goals (1-6) to locate and reach the human. (center) Robot's observations upon reaching each global goal (1-6) during its exploration of the environment. (right) Robot's final observation upon successfully reaching the human at an appropriate angle, depending on the human's activity, to initiate a conversation.}
  \label{fig:HRI_example}
 \end{figure*}

Robots that can autonomously navigate and interact with humans hold immense potential in various assistive scenarios. Consider a robot assisting an elderly individual in their home, helping with daily activities like cooking, medication reminders, or encouraging physical activity. However, to fulfill these tasks effectively, robots must first locate and approach humans appropriately. Our research addresses this crucial challenge in embodied visual navigation, where an agent utilizes visual input to navigate complex 3D environments, avoid obstacles, and reach desired destinations. An example of such a complex task is shown in Fig \ref{fig:HRI_example}.
 
Over the past decade, significant advancements in this field have been driven by the availability of photorealistic 3D scene datasets \cite{proceeding5,proceeding6,proceeding7} and fast navigation simulators \cite{proceeding4,proceeding7,proceeding8}. Our work builds on these advancements but takes a systematic approach to investigate the task of locating and navigating to a human for initiating interactions. We introduce a pipeline that leverages global and local goal policies to efficiently find and reach humans in complex environments with a focus on assistive tasks and enhancing human-robot interactions enhancement. To achieve this we utilize the Habitat AI simulator \cite{proceeding4} and the Gibson dataset \cite{proceeding7} for this task conducting a comparison with different baseline methods. Our aim is to evaluate how well the robot performs in finding and navigating toward humans who are engaged in activities such as cooking, eating, being, on a call, or watching TV. In each scenario, the robot needs to explore the environment in order to find the human and approach them safely. In our study, we present a well-defined problem formulation and establish a set of baseline methods that integrate human detection and point-goal navigation to address this challenge effectively.

\subsubsection{PH-Ego}


\begin{figure*}[!htpb]
\centering
\includegraphics[width=\textwidth]{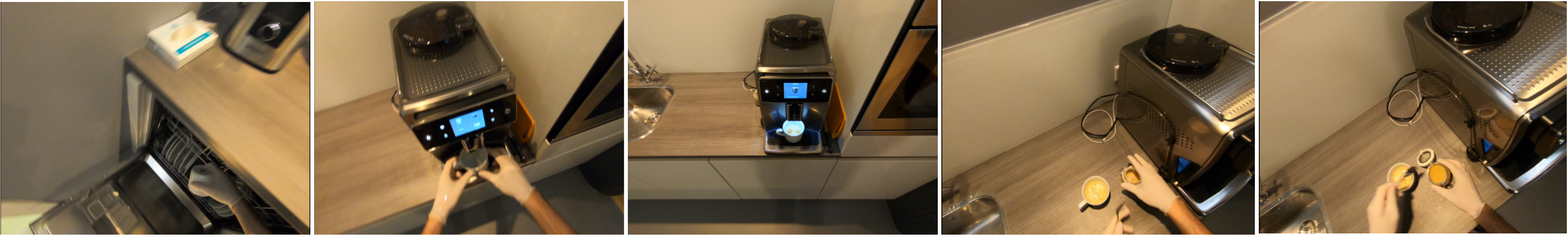}
\caption{The example sequence of frames from PH-Ego dataset capturing coffee preparation and adding sugar actions.}
\label{fig:phego-example}
\end{figure*}

The egocentric action anticipation task consists in predicting future (unobserved) actions performed by a camera-wearer based on input video from a wearable first-person view camera. We are focusing on applying egocentric action anticipation methods to the personal health domain, i.e., utilizing them for the analysis of dietary and hygienic activities routine and for the prevention of undesirable behavior (such as tasting food before washing hands or adding sugar). We collect a dataset PH-Ego, consisting of egocentric videos, capturing specific activities related to food preparation; fine-tune existing action anticipation models on this dataset and analyse effects caused by domain shift. 

The dataset was collected from nine healthy volunteers residing in Eindhoven, Netherlands, during the summer of 2022. To maintain privacy and consistency, participants were instructed to remove any hand accessories and wear latex gloves during the video recording process.

The data collection was facilitated using a GoPro Hero 7 Black camera, mounted on the head of each participant. The videos were captured at a frame rate of 50 fps and a resolution of 1920x1280 pixels, with each video having an average duration of 6.5 minutes. In total, 67 action segments were recorded.

For the purposes of model training and validation, the videos of seven participants were utilized in the training dataset, and the videos of the remaining two participants were reserved for model validation. An illustrative example of a sequence of frames, showcasing the actions of \textit{making coffee} and \textit{adding sugar}, is depicted in Figure~\ref{fig:phego-example}.

\subsubsection{Imagym}

For the last image-based task, we utilize a dataset of fetal ultrosound images to propose a novel data-driven patient simulator for decision support and automation in the field of obstetric ultrasonography \cite{obstetrics-sonography}.

The goal of this simulator is to accurately model the job of an ultrasound sonographer in the context of a patient undergoing a pregnancy in a way compatible with modern Reinforcement Learning methods \cite{liDeepReinforcementLearning2017} to pave the way for autonomous or semi-autonomous ultrasound sonography \cite{autonomous-ultrasound-review}.
The job in question entails moving the ultrasound probe along the patient's body in order to aquire an image of the fetus that satisfies the guidelines for fetal-screening \cite{isoug-guidelines}, most importantly, the fact that the fetus' stomach and umbilical vein are on the image while their heart is not.

\begin{figure*}
    \centering
    \begin{subfigure}{.45\linewidth}
      \centering
      \includegraphics[width=.95\linewidth]{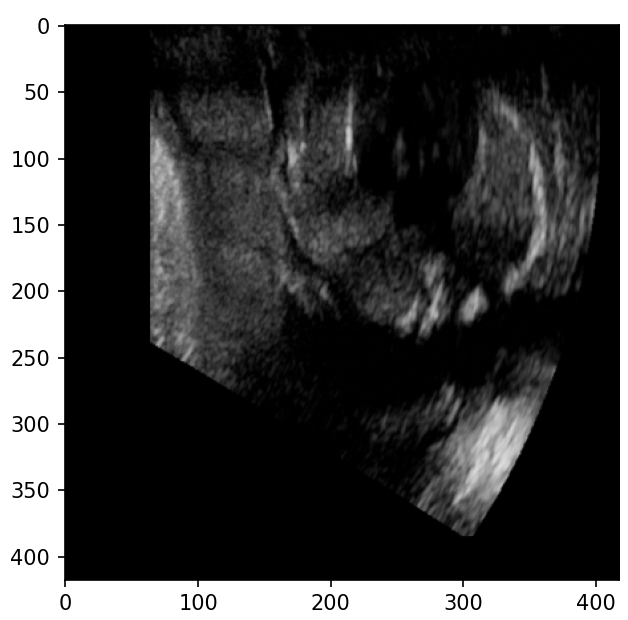}
      \caption{Starting point: the probe is located at the exact average of available positions. Heart, stomach and umbilical vein are unseen.}
      \label{fig:img-before}
    \end{subfigure}%
    \begin{subfigure}{.45\linewidth}
      \centering
      \includegraphics[width=.95\linewidth]{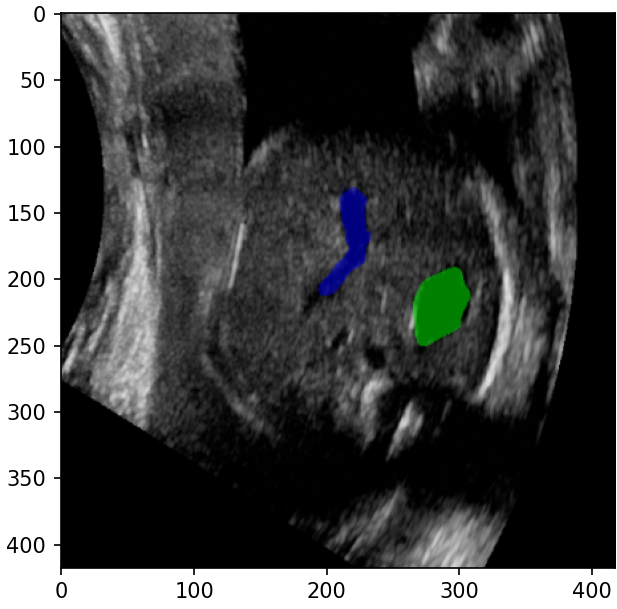}
      \caption{Endpoint: stomach (green) and umbilical vein (blue) are present, heart is absent.}
      \label{fig:img-after}
    \end{subfigure}
    \caption{Two examples of the agent's observation at different positions of the probe}
    \label{fig:imgs}
\end{figure*}

The simulator is based on a dataset of 3D volumes representing fetal abdominal ultrasound scans (though be easily adapted to other medical imaging scenarios as long as a dataset of volumes and relevant organ annotations is available).
Each scan is a function defined over a cuboid $B$ of size $\langle x_\text{max},y_\text{max},z_\text{max} \rangle$ (in millimeters).
\begin{equation}
    I(x,y,z): B \rightarrow [0;1]
\end{equation}

Each scan is accompanied by three mask images $M_\text{heart}$, $M_\text{stomach}$ and $M_\text{uv}$ over the same domain, indicating which parts of the scan are considered to be the heart, the stomach and the umbilical vein respectively.

\begin{equation}
    M(x,y,z): B \rightarrow \{0,1\}
\end{equation}

The state, in this case, is the current position of the proble defined by 2 3-vectors, the position of the probe $s_\text{loc} \in \mathcal{S}_\text{loc}$ and its direction $s_\text{dir} \in R^{3}$.

The simulator has 2 modes for the space of possible locations $\mathcal{S}_\text{loc}$: \emph{free} and \emph{realistic} movement.
In \emph{free movement} mode, $\mathcal{S}_\text{loc} = B$, thus the agent can place the probe anywhere within the bounds of the image, even inside the patient.
In \emph{realistic movement} mode, $\mathcal{S}_\text{loc} \subset B$ representing the surface of the patient's body available to the agent.

The action space $\mathcal{A} = R^{7}$, where any $a \in \mathcal{A}$ can be decomposed as

\begin{equation}
    a = (a_{x} , a_{y} , a_{z} , a_\text{roll} , a_\text{pitch} , a_\text{yaw}, a_\text{end})
\end{equation}

The first 3 components modify the $s_\text{loc}$, the next 3 modify $s_\text{dir}$ and, finally if $a_\text{end} > 0$, the episode terminates and the current image is considered final.
Note that in \emph{realistic movement} mode, blind application of $(a_{x} , a_{y} , a_{z})$ can lead to the probe being placed outside of the allowed domain of $\mathcal{S}_\text{loc}$.
In this case, a legal location will be chosen, in a manner that minimizes the distance between requested and real probe position:

\begin{equation}
    s_\text{loc} \leftarrow \min_{s \in \mathcal{S}_\text{loc}} \left\{ \lVert s - (s_\text{loc} + (a_{x} , a_{y} , a_{z})) \rVert \right\}
\end{equation}

The agent's observation is obtained by projecting $I(x,y,z)$ onto a plane defined by the current position of the probe $s_\text{loc}$ and its direction $s_\text{dir}$, simulating the image that the sonographer would see on their screen.
It is, theoretically, a function $I_\text{proj}(x_\text{proj}, y_\text{proj})$, however, since it is customary in POMDP framework for observations $o$ to be matrices, we replace it with a matrix of evaluations of $I_\text{proj}(x_\text{proj}, y_\text{proj})$ on an arbitrary (hyperparametrized) coordinate grid $\{(x_\text{proj}, y_\text{proj})\}$.

The quality metric of the image is defined in accordance with ISOUG guidelines \cite{isoug-guidelines} to be measured by the surface area of the heart, the stomach and the umbilical vein on the projection, normalized by the volume of these organs.

\begin{equation}
    Q(I_\text{proj}) = \frac{S_\text{stomach}(I_\text{proj})}{V_\text{stomach}(I)} + \frac{S_\text{heart}(I_\text{proj})}{V_\text{heart}(I)} + \frac{S_\text{uv}(I_\text{proj})}{V_\text{uv}(I)}
\end{equation}

The reward assigned to the agent then is the difference in quality between the current image and the one obtained at the previous iteration, so that probe movements that improve the image are rewarded positively and all rewards obtained during the episode sum up to the quality of the chosen image.

\subsection{Natural language perspective}

Many chronic health conditions such as diabetes, obesity, substance abuse, and sleep disorders, can be helped by a lifestyle change. Passive health self-management services with a watch and a tracking app do not seem to be very effective for lifestyle change~\cite{finkelstein_effectiveness_2016,wise_activity_2016}. Health counseling, on the other hand, is known to work~\cite{rubak_motivational_2005,hofmann_efficacy_2012}. However, health counseling is expensive and do not scale well in a world with a growing deficit of healthcare workers. One solution for personal health self-management is automated counseling, AC. After 50 years of research, see, e.g., \cite{bickmore_health_2006}, the recent progress in dialog system technologies is finally making AC a realistic option. AC has also recently been shown to be effective in a controlled trial \cite{fitzpatrick_delivering_2017}.
%
%
%
%
%
\begin{figure}[htp!]
\begin{center}
    \includegraphics[width=1\linewidth]{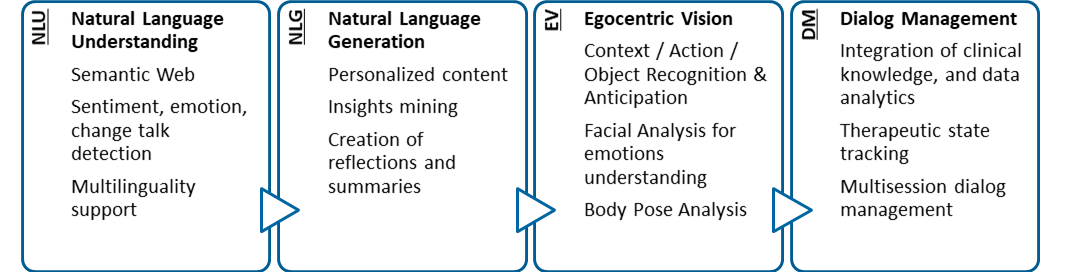}
	\caption{Challenges in conversations technologies and applications for health self-management.}
	\label{trends}
\end{center}
\end{figure}

We believe that collaborative care management with a conversational agent will be central for future health self-management. The key element is the engaging dialogue, which makes the user reflect own lifestyle choices and barriers and find opportunities and motivation for a change~\cite{miller_toward_2009}. This requires advances in sensing technology, model-based cognitive interaction technologies, and embodiments. Figure~\ref{trends} lists enabling technologies that we believe are central for conversational interfaces in health self-management, but where also breakthroughs are needed. 

Users may have a very different lifestyle and requirements, which requires highly adaptive solutions. In NLU, NLG, and dialog management~\cite{zora,fred} there is a clear progression towards deep E2E solutions which provide flexibility over conventional rule-based systems~\cite{constantin_end--end_2018,gehrmann_end--end_2018,madotto_mem2seq:_2018}. However, in healthcare, it is very important to be able to combine data-driven learning with validated clinical knowledge~\cite{RiboniBCJH16}, which is a challenge for many deep-learning solutions. The system must understand what the words of the user mean in the desired clinical and therapeutic setting~\cite{DBLP:conf/ijcai/ChaouaRCHH18, ChaouaRCHH182}, and how the output is expected to influence the user, and the goals of the counseling.

In order to lead a meaningful conversation about the lifestyle of the user, the system has to have a good understanding of it. This requires data-driven techniques for modeling of the lifestyle \cite{harma_probabilistic_2016,van_wissen_optimization_2016}, and situational intelligence, which can put the conversation in the right context~\cite{geriatric}. In particular, wearable sensors and {\it egocentric} vision systems~\cite{Kanade2012}, together with cameras integrated into the dialog agent, seem to be necessary for human-level situational intelligence. Wearable computing has a long history~\cite{Betancourt2015,Mann97}, it has become of more interest in the last years with the advancement of both hardware and software technologies~\cite{Furnari2017a}. The main applications of FPV systems in the context of assistive Computer Vision~\cite{Leo2017} is related to memory augmentation~\cite{Damen2015} and life logging~\cite{Furnari2018b,Gurrin2014,Ortis2017}. Recent studies have also considered the FPV paradigm to recognize important objects observed by user~\cite{Lee2015} and to understand or to anticipate the actions and activities performed by the user~\cite{Damen2018EPIC,Furnari2018a,Ryoo2016,Singh2016} or the next object to which the user is going to interact with~\cite{Furnari2017c}. 
A robotic embodiment for the conversational agent is another way to extend the capabilities of the agent. Robotic embodiments have proven to attract more interest and engagement from patients rather than other technologies. Different studies~\cite{mann2015people, feingold2018differences} reported that both old and young adults preferred to interact with embodied robot over the non-embodied computer screen, and thus carebots may offer benefits over smartphones or tablets in delivering healthcare. 

In recent years, Large Language Models (LLM) have successfully imparted close to human-like abilities for several NLP downstream tasks~\cite{devlin-etal-2019-bert,li2019fine,rogers2020primer,budzianowski2019hello,radford2018improving}. However, there are two significant constraints: a) available training data size~\cite{sdaih23,kumar2023data} and b) the complex domain-specific context jeopardizing LLM's optimal and reliable performance~\cite{ DBLP:conf/iui/DessiHKRR20,9286431}. At first, it may seem that "available training data size' should not be a problem at all due to the vast amount of data humanity produces. However, in reality, usable data in several domains, including healthcare, is scarce and not publicly available to train deep neural network models sufficiently. 

A potential solution to these limitations is to equip the LMs with domain. While the available research works use Knowledge Graphs Embeddings (KGEs)~\cite{wang2017knowledge} to inject domain knowledge~\cite{bordes2013translating,wang2014knowledge,lin2015learning,schlichtkrull2018modeling,trouillon2016complex,sun2019rotate,nickel2016holographic,shang2019end,nguyen2018novel,xie2016representation}. KGE has proven to be helpful in incorporating world knowledge, but it is still debatable if the KGEs sufficiently capture KGs semantics~\cite{jain2021embeddings}. The work in~\cite{9866735} provides a Knowledge-based LLM to use the Resource Description Framework (RDF) triples directly at the input level. The proposed model K-LM works at the crossroads of Generative Pretrained Transformer (GPT-2) and Bidirectional Encoder Representations from Transformers (BERT). K-LM uses a novel pipeline to select, categorize, and filter the RDF triples and introduces heuristic methods to inject domain-specific Knowledge into Knowledge LM.

\subsubsection{Diet coaching}


NHS UK defines malnutrition as the situation caused by either undernutrition or overnutrition~\cite{nhs-malnutrition}. Nowadays, malnutrition has become a major public health issue~\cite{diet-stats-worldwide}, with profound consequences in society. This problem is particularly acute in UK~\cite{bma-diet,ultraprocessed-food-survey-eu} where 63\% of the UK adult population are overweight, and 27\% are obese. Undernutrition is estimated to affect three million people in UK~\cite{bda-malnutrition}, especially among elders and socially isolated individuals.
To this regard, obesity is a known cause of multiple illnesses including Type 2 diabetes, infertility, cancer and cardiovascular diseases~\cite{obesity-act-scot}; undernutrition leads to unplanned weight loss, reduced energy, infections and more~\cite{nhs-malnutrition}. The economic impact of unhealthy eating is notable: every year, obesity-related diseases put £6bn of additional costs on NHS, impacting wider society for about £27bn~\cite{bma-diet}. These costs are expected to reach £9.7bn and £49.9bn by 2050 respectively. Undernutrition is estimated to cost £19bn in England alone~\cite{bda-malnutrition}.

Taken together, the effects of malnutrition on people's health and on countries' economies shed light on the need for innovative technologies to encourage healthy eating. We contribute to this by presenting a series of tailored and understandable communication technologies for diet coaching, with a focus on the use of Natural Language Generation (NLG), that is the set of technologies capable of producing text from structured data representation~\cite{reiter-dale-2000,puzikov2018e2e}. In details, we implement and evaluate both static and dynamic NLG systems to deliver personalised diet reports.  

\subsubsection{AnnoMI}


Motivational interviewing (MI,~\cite{mi}) is an important counselling style aimed at evoking a client's own motivation to adopt a certain positive behaviour change, such as smoking cessation and alcohol use reduction. As a result, MI is widely used in healthcare-related scenarios~\cite{mi-in-healthcare}.

Natural language processing (NLP) for MI has seen considerable progress in recent years (e.g.,~\cite{text-based-behaviour-coding-forecasting}), but it has been substantially limited by the lack of available MI dialogue data and annotations, which is mostly attributable to the concerns and regulations w.r.t. client confidentiality and data sharing.

To address this challenge, we present AnnoMI (\cite{annomi-icassp,annomi-journal}), the first publicly available expert-annotated dataset of MI dialogues. We sourced the conversations of AnnoMI from professionally produced therapy demonstration videos, where each video is a one-to-one dialogue between a therapist and a client. Those conversations cover a wide range of topics.

Notably, AnnoMI is unambiguously legal and well-transcribed, as we 1) gained explicit consent from the video owners for the creation and release of the dataset; 2) opted for a professional transcription service to obtain high-quality transcripts.

AnnoMI also comes with utterance-level annotations from professional therapists. Two major annotated attributes are \textit{therapist (main) behaviour} for therapist utterances and \textit{client talk type} for client utterances: \textit{therapist (main) behaviour} indicates the (main) action/behaviour displayed in a therapist utterance, while \textit{client talk type} shows whether the client is leaning towards, showing resistance to, or not showing any particular incliniation w.r.t. the behaviour change goal.

\subsubsection{Insight mining}


Insights from personal health data help the population make better lifestyle choices leading to better health and well-being. The Cambridge Dictionary defines an insight as "a clear, deep, and sometimes sudden understanding of a complicated problem or situation, or the ability to have such an understanding". In the domain of data science, it is an overloaded term \cite{law2020characterizing} used in many perspectives such as findings obtained from data \cite{saraiya2005insight} or the outcome of evaluating data findings with domain knowledge \cite{sacha2014knowledge}. In line with the earlier perspective, we define insight as statements that help us understand patterns or behaviors observable from data. Behaviors that stand out, for example, "You sleep less on Mondays than other days", are worth communicating to the user with the hope of a remedy. 
Therefore, we introduced the insight generator framework \cite{susaiyah2020towards} that automatically generates such insights and picks the most useful ones to show to the user. The generation of such insights requires us to define an abstract structure of such insights that we refer to as an insight schema \cite{susaiyah2022aberdeen}. For example, "You {{measurement}} {{comparison}} {{time\_period1}} {{time\_period2}}" is an insight-schema that defines all insights that compare a measurement such as sleep duration between two time periods such as Mondays and other days.
The framework iteratively checks through the data looking for insights matching the schema and generates insights candidates when a match is found. The candidates are later ranked based on various factors such as significance, accuracy, user preference, and reliability \cite{susaiyah2021neural,susaiyah2022aberdeen,susaiyah2022RL,ecai-paper}. Finally, the top-ranking insights are recommended to the user.

\section{Evaluations}

\subsection{MIMIC-IV-Ext-SEQ}

We train a baseline model that consists of a two-layer multilayer perceptron (MLP) with 1000 hidden layer size RELU \cite{agarapDeepLearningUsing2018} activation function and batch normalization after each layer. As input, all events from the first day are used and one-hot encoded in a 87899-dimensional vector. As prediction target, all events from the second day are used and encoded via their clustering mapping, e.g. c10, c100, c1000, c10000, into a vector of the corresponding dimension. As objective function we used binary cross entropy. The last layer contains a sigmoid function which transforms the output to probabilities for each vector element. A threshold is set at 0.5 to decide if an event occurs or not. All models were trained with batch size 512 for 3 epochs.

We test our baseline on the \emph{second day event prediction task} and summarize the results for different clusterings in table \ref{tab:mytable2}. It can be seen that the more classes are in the clustering, the harder the prediction task becomes. As noted above, accuracy is a deceptive metric in this scenario.

\begin{table}[h]
  \centering
    \begin{tabular}{lcccc} \toprule
        {clustering} & {recall} & {accuracy} & {precision} & {F1}  \\ \midrule
        {c10}  & 0.903 & 0.840 & 0.790 & 0.827 \\
        {c100}  & 0.568 & 0.885 & 0.713 & 0.632 \\
        {c1000}  & 0.500  & 0.976 & 0.710  & 0.586 \\
        {c10000}  & 0.509  & 0.996 &  0.703  & 0.589 \\ \midrule
    \end{tabular}
  \caption{Evaluation results of 2 x 1000 MLP for first day - second day  prediction, entire dataset}
  \label{tab:mytable2}
\end{table}

For many patients it is the case that they are in the hospital for only one day. For these patients, the model should predict no event on the second day. However, evaluation of the models showed that this is almost never the case. However, one can argue that it is more important to get problematic patients correct than the ones who leave the hospital after one day. Therefore, we evaluated the model also only on patients which are in the hospital for at least 2 days. Since the data is roughly ordered according to length of stay, we achieved this by skipping the first 100k samples in the training and the first 1k samples in the test data. The models shows improved performance here, as can be seen in Table \ref{tab:mytable1}.

\begin{table}[h]
  \centering
    \begin{tabular}{lcccc} \toprule
        {clustering} & {recall} & {accuracy} & {precision} & {F1}  \\ \midrule
        {c10}  & 0.942  & 0.854 & 0.850  &  0.890 \\ 
        {c100}  & 0.633  & 0.878 & 0.765  &  0.692 \\ 
        {c1000}  & 0.548  & 0.974 & 0.770  &  0.640 \\ 
        {c10000}  &  0.539  & 0.995 & 0.771  &  0.634 \\ \midrule
        
    \end{tabular}
  \caption{Evaluation results of 2 x 1000 MLP for first day - second day  prediction, skipping first 100k train / 1k test samples}  \label{tab:mytable1}
\end{table}


The same setup but with 3 hidden layers and 5000 units each shows improved performance as can be seen in table \ref{tab:mytable3}. Bigger models were tested as well, but showed no further improvement.

\begin{table}[h]
  \centering
    \begin{tabular}{lcccc} \toprule
        {configuration} & {recall} & {accuracy} & {precision} & {F1}  \\ \midrule
        {1}  & 0.505  & 0.996 &  0.727  & 0.595 \\
        {2} & 0.544  & 0.9959 &  0.784  & 0.642 \\  \midrule
    \end{tabular}
  \caption{All evaluations with c10000; 1: Evaluation results of 3 x 5000 MLP for first day - second day  prediction, entire dataset; 2: Evaluation results of 3 x 5000 MLP for first day - second day prediction, skip first 100k train / 1k test samples}
  \label{tab:mytable3}
\end{table}

Replacing ones in the one-hot encoding with the corresponding events' intensities impaired the models' performance, likely because it introduces a false equivalency between a zero-intensity event and lack of an event, i.e. "average blood pressure" is different from "no blood pressure measurement".

\subsection{Auto-ALS}

Two experimental emergency care agents were train with Reinforcement Learning in  \cite{bhalwankarNeuroSymbolicReinforcementLearning}: one used a common Proximal Policy Optimization (PPO) algorithm and no prior knowledge of emergency care, another implemented several common sense constraints on the agent in Linear Temporal Logic and trained a Graph Neural Network that took those constraints into account.
See table \ref{tab:auto-als-params} for details of the agents.

\begin{table}[]
    \centering
    \begin{tabular}{|l|l|l|}
    \hline & PPO & GNNprog w/rules \\
    \hline Env. steps per update  & 2048 & 2048 \\
    \hline Number of epochs & 4 & 4 \\
    \hline Minibatch Size  & 256 & 1024 \\
    \hline Discount factor ()  & 0.99 & 0.99 \\
    \hline Learning rate  & 0.0001 & 0.0001 \\
    \hline GAE-  & 0.95 & 0.95 \\
    \hline Entropy coefficient  & 0.01 & 0.01 \\
    \hline Value loss coefficient  & 0.5 & 0.5 \\
    \hline Gradient Clipping  & 0.5 & 0.5 \\
    \hline PPO Clipping ()  & 0.2 & 0.2 \\
    \hline
    \end{tabular}
    \caption{Hyperparameters of evaluated agents}
    \label{tab:auto-als-params}
\end{table}

The results are presented in table \ref{tab:auto-als-stats}

\begin{table}[]
    \centering
    \begin{tabular}{|l|l|l|l|l|l|}
    \hline \multicolumn{6}{|c|}{ Group Statistics } \\
    \hline & $\mathrm{v} 26$ & $\mathrm{~N}$ & Mean & \begin{tabular}{l} 
    Std. \\
    Deviation
    \end{tabular} & \begin{tabular}{l} 
    Std. Error \\
    Mean
    \end{tabular} \\
    \hline \begin{tabular}{l} 
    average \\
    discounted \\
    return
    \end{tabular} & 1 & 781 & 0.85 & 1.02 & 0.04 \\
    \cline { 2 - 6 } & 2 & 781 & -4.04 & 7.03 & 0.25 \\
    \hline
    \end{tabular}
    \caption{Group Statistics w.r.t to average discounted returns on LTL vs No-LTL model outputs in AutoALS Env}
    \label{tab:auto-als-stats}
\end{table}

\subsection{Human-Robot interaction}


We evaluate the robot's performance in locating and navigating to humans engaged in various activities, such as cooking, eating, on a call, and watching TV. We validate the feasibility of this task and the effectiveness of our proposed baselines using the Habitat AI simulator \cite{proceeding4} and the Gibson modified dataset, which consists of twenty complex 3D environments. Since the Gibson dataset lacks human models, we have created human models in poses aligned with different activities (eating, cooking, watching TV, and on a call). These human models were strategically placed in various locations within the Gibson environments, such as the kitchen, TV lounge, bedroom, and other relevant areas.

We run $2000$ evaluation episodes, with each scene containing $100$ episodes. We consider the two variants of the task: V1 aims to reach the human from any angle, whereas in V2 episode success depends on the orientation difference between the robot and the human at the end of the episode. Table \ref{results} provides Success weighted by Path Length (SPL) and success rate (SR) for both V1 and V2. Our proposed approach achieves $93\%$ SR and $57\%$ SPL under $V1$. However, in $V2 (\theta=30^{\circ})$, our approach only achieved $26\%$ SR and $14\%$ SPL. This suggests that V2 of the task is much more challenging and more research is still needed. Overall, the considered baselines achieve promising performance in locating and reaching humans in previously unseen environments. However, further research is required in this exciting field. We believe that our proposed approach serves as a first step towards building autonomous task-oriented assistive robots for home use cases.

\begin{table*}[t]
\caption{Average SPL and SR on the Gibson validation dataset.}\label{results}
\centering
\begin{tabular}{l@{\hspace{1.0cm}}c@{\hspace{1.0cm}}c@{\hspace{1.0cm}}c}
\hline
 Task &  SPL ($\uparrow$) & SR ($\uparrow$)\\
\hline
V1 (any angle) & $0.57$&$0.93$ \\

V2 ($\theta \leq 60^{\circ}$) & $0.25$&$0.44$\\
V2 ($\theta \leq 30^{\circ}$) & $0.14$&$0.26$\\
\hline

\end{tabular}
\end{table*}

\subsection{PH-Ego}


The task of \textit{trimmed action anticipation} involves predicting the class $y$ of an impending action, initiating at time $\tau_s$, by analyzing a trimmed video segment that precedes the action. The final observed timestamp is defined as $\tau_s-\tau_a$, where $\tau_a$ represents the anticipation time, typically established at 1 second \cite{damen2022rescaling}.

In contrast, the \textit{untrimmed action anticipation} task operates without reliance on pre-annotated action annotations, executing anticipations and their subsequent evaluations at intervals of every $\alpha$ seconds, irrespective of whether an action is forthcoming or not \cite{rodin2022untrimmed}. This untrimmed approach holds significant relevance in the context of personal assistant settings, where the system necessitates continuous monitoring of activities to facilitate timely alerts or assistance. However, this method also carries an elevated risk of generating false-positive predictions.

In the study \cite{rodin2022untrimmed}, we propose the utilization of the mean Average Precision (mAP) as an evaluation metric for untrimmed action anticipation. Drawing inspiration from object detection benchmarks, the research suggests treating untrimmed anticipation as a future action detection problem. This approach encourages the diversification of models, factoring in false-positive predictions to enhance the robustness and accuracy of the anticipatory models in identifying and forecasting future actions. However, for a smaller problem, where the time-to-action predictions do not play important roles, we propose to use simpler classification metrics \cite{rodin2023egocentric}.

\subsection{AnnoMI}


Based on the publicly available AnnoMI dataset that we created, we explore two natural language understanding (NLU) tasks with real-world applicability: 1) current-turn therapist/client behaviour prediction (\textbf{prediction task},~\cite{annomi-journal}); and 2) next-turn therapist behaviour forecasting (\textbf{forecasting task},~\cite{mi-action-forecasting}). The code and data of our experiments are publicly available.

The prediction task aims to predict the (main) behaviour of each therapist utterance and the talk type of each client utterance, given the utterance itself as the input. An accurate prediction model would be readily usable for automatically annotating transcripts of MI sessions, which would considerably save resources on laborious human annotation and accelerate insight generation from recorded sessions.

Overall, we found BERT with adapters~\cite{houlsby-adapters}~\textemdash~a parameter-efficient version of BERT~\cite{devlinBERTPretrainingDeep2019}~\textemdash~to be the best-performing model, with performance higher for therapist behaviours than for client talk types. Encouragingly, for therapist behaviour prediction, the model also proved to maintain a high level of performance when tested on utterances from new dialogue topics to which it had not been exposed during training, which is meaningful for real-world deployment.

The forecasting task aims to forecast which (main) behaviour a good therapist would show in their next-turn response to the client, given the history of the ongoing dialogue where the last turn comes from the client.  An accurate forecasting model would be able to assist a (junior) therapist in real time in deciding what strategy to choose for responding to the client.

For this task, we explored a variety of NLP modelling choices with the RoBERTa~\cite{roberta} model, but the performance was suboptimal. We believe that this reflects the considerable flexibility in how a good therapist could respond to a client, as there are often multiple ``optimal behaviours/actions'' to display in the next turn. Therefore, future work could model this task in a probabilistic setup to accommodate this flexibility.

The information retrieval task aims to filter therapeutical content by estimating quality automatically~\cite{10.1007/978-3-031-37249-0_10}. First, we applied state-of-the-art information retrieval models to evaluate their applicability in the psychological domain for ranking therapy sessions by estimated quality. Then, given the sensitive psychological information associated with each therapy session, we analyzed the potential risk of unfair outcomes across therapy topics, i.e., mental issues, under a common fairness definition. Our experimental results show that the employed ranking models are reliable for systematically ranking high-quality content above low-quality one, while unfair outcomes across topics are model-dependent and associated low-quality content distribution.

\subsection{Insight mining}


Insights are subjective and evaluating them requires human participants. Therefore, Evaluating Insight Generation systems is time-consuming and expensive due to recruitment, compensation, sample size, etc. Therefore, we examined the possibility of using Reinforcement Learning (RL) to build and Evaluate Insight mining systems \cite{susaiyah2022RL}. Here, we first developed a simulator that generates a human lifestyle that assumes performing a certain task at a given time of the day, for example, waking up, having breakfast, communicating to work, etc. We then set up an RL environment that iteratively trains an agent to recommend useful insights to the simulated user. The insights modify the lifestyle of the user and this leads to different insights and vice versa. The cost function is determined by the lifestyle parameter that is being improved. In our experiments, we used the Pittsburg sleep quality index (PSQI) as a reward function. Experiments show that the above approach was able to recommend insights that significantly quickened the adaptation of a behavior change towards a better lifestyle in the simulated environment compared to baseline techniques.

\section{Discussion}

Our benchmark suite demonstrates the state of the art across different tasks of Machine Learning in Healthcare. All problems discussed are, however, open and PhilHumans is meant to be an evaluation tools for future methodological research.

\section{Acknowledgements}
\label{sec:acb}
This project has received funding from the European Union’s Horizon 2020 research and innovation programme under grant agreement No 812882.

\newpage
\bibliographystyle{spmpsci}
\bibliography{health_dialogue,vadim,alex,allmin,asfand, simone,ivan, vivek}

\begin{thebibliography}{100}
\providecommand{\url}[1]{{#1}}
\providecommand{\urlprefix}{URL }
\expandafter\ifx\csname urlstyle\endcsname\relax
  \providecommand{\doi}[1]{DOI~\discretionary{}{}{}#1}\else
  \providecommand{\doi}{DOI~\discretionary{}{}{}\begingroup
  \urlstyle{rm}\Url}\fi

\bibitem{SurveyShowsHidden1993}
Survey shows hidden cuts.
\newblock Nursing Standard \textbf{7}(35), 23--24 (1993).
\newblock Publisher: RCN Publishing Ltd.

\bibitem{UnderstaffingSignificantIssue2012}
Understaffing is ‘significant issue’ in nursing homes.
\newblock Nursing Standard \textbf{27}(13), 6--6 (2012).
\newblock Publisher: RCN Publishing Ltd.

\bibitem{diet-stats-worldwide}
Afshin, A., Sur, P.J., Fay, K.A., Cornaby, L., Ferrara, G., Salama, J.S.,
  Mullany, E.C., Abate, K.H., Abbafati, C., Abebe, Z., et~al.: Health effects
  of dietary risks in 195 countries, 1990--2017: a systematic analysis for the
  global burden of disease study 2017.
\newblock The Lancet \textbf{393}(10184), 1958--1972 (2019)

\bibitem{agarapDeepLearningUsing2018}
Agarap, A.F.: Deep learning using rectified linear units (relu).
\newblock arXiv preprint arXiv:1803.08375  (2018)

\bibitem{agrawalMachineLearningHealthcare2020}
Agrawal, R., Chatterjee, J.M., Kumar, A., Rathore, P.S., Le, D.N.: Machine
  {Learning} for {Healthcare}.
\newblock Chapman and Hall/CRC (2020)

\bibitem{Anshik2021Handling}
{Anshik}: Handling {Availability} of {Low}-{Training} {Data} in {Healthcare}.
\newblock In: {AI} for {Healthcare} with {Keras} and {Tensorflow} 2.0, pp.
  173--214. Apress (2021)

\bibitem{ashleyy.metcalfHospitalUnitUnderstaffing2016}
{Ashley Y. Metcalf}: Hospital {Unit} {Understaffing} and {Missed} {Treatments}:
  {The} {Moderating} {Effect} of {Teamwork}.  (2016)

\bibitem{geriatric}
Asprino, L., Gangemi, A., Nuzzolese, A.G., Presutti, V., Recupero, D.R., Russo,
  A.: Autonomous comprehensive geriatric assessment.
\newblock In: Proceedings of the 1st International Workshop on Application of
  Semantic Web technologies in Robotics co-located with 14th Extended Semantic
  Web Conference {(ESWC} 2017), Portoroz, Slovenia, May 29th, 2017., pp. 42--45
  (2017).
\newblock \urlprefix\url{http://ceur-ws.org/Vol-1935/paper-05.pdf}

\bibitem{zora}
Atzeni, M., Recupero, D.R.: Deep learning and sentiment analysis for
  human-robot interaction.
\newblock In: The Semantic Web: {ESWC} 2018 Satellite Events - {ESWC} 2018
  Satellite Events, Heraklion, Crete, Greece, June 3-7, 2018, Revised Selected
  Papers, pp. 14--18 (2018).
\newblock \doi{10.1007/978-3-319-98192-5\_3}.
\newblock \urlprefix\url{https://doi.org/10.1007/978-3-319-98192-5\_3}

\bibitem{bda-malnutrition}
BDA: Spotting and treating malnutrition: Food fact sheet.
\newblock \url{https://www.bda.uk.com/resource/malnutrition.html} (2022).
\newblock Accessed on 7/12/2022

\bibitem{Betancourt2015}
Betancourt, A., Morerio, P., Regazzoni, C.S., Rauterberg, M.: The evolution of
  first person vision methods: A survey.
\newblock IEEE Transactions on Circuits and Systems for Video Technology
  \textbf{25}(5), 744--760 (2015).
\newblock \doi{10.1109/TCSVT.2015.2409731}

\bibitem{bhalwankarNeuroSymbolicReinforcementLearning}
Bhalwankar, R.: Neuro-{Symbolic} {Reinforcement} {Learning} for {Sepsis} and
  {Emergency} {Care} {Treatment}

\bibitem{bickmore_health_2006}
Bickmore, T., Giorgino, T.: Health dialog systems for patients and consumers.
\newblock Journal of Biomedical Informatics \textbf{39}(5), 556--571 (2006).
\newblock \doi{10.1016/j.jbi.2005.12.004}

\bibitem{bma-diet}
BMA: Improving the nation’s diet: action for a healthier future.
\newblock
  \url{https://www.bma.org.uk/what-we-do/population-health/supporting-people-to-live-healthier-lives/improving-the-nation-s-diet-action-for-a-healthier-future}
  (2021).
\newblock Accessed on 10/11/2022

\bibitem{bordes2013translating}
Bordes, A., Usunier, N., Garcia-Duran, A., Weston, J., Yakhnenko, O.:
  Translating embeddings for modeling multi-relational data.
\newblock Advances in neural information processing systems \textbf{26} (2013)

\bibitem{susaiyah2022RL}
Braz, L.G., Susaiyah, A.P.S., Petkovic, M., H{\"a}rm{\"a}, A.: Deep
  reinforcement learning based insight selection policy  (2022)

\bibitem{briskAIEnhanceInteractive2018}
Brisk, R., Bond, R.R., Liu, J., Finlay, D., McLaughlin, J., McEneaney, D.: {AI}
  to enhance interactive simulation-based training in resuscitation medicine.
\newblock In: British {HCI} {Conference} 2018 (2018)

\bibitem{brown2020:language}
Brown, T.B., Mann, B., Ryder, N., Subbiah, M., Kaplan, J., Dhariwal, P.,
  Neelakantan, A., Shyam, P., Sastry, G., Askell, A., Agarwal, S.,
  Herbert-Voss, A., Krueger, G., Henighan, T., Child, R., Ramesh, A., Ziegler,
  D.M., Wu, J., Winter, C., Hesse, C., Chen, M., Sigler, E., Litwin, M., Gray,
  S., Chess, B., Clark, J., Berner, C., McCandlish, S., Radford, A., Sutskever,
  I., Amodei, D.: Language models are few-shot learners.
\newblock In: International {Conference} on {Neural} {Information} {Processing}
  {Systems}, {NIPS}'20, pp. 1877--1901. Curran Associates Inc., Red Hook, NY,
  USA (2020)

\bibitem{budzianowski2019hello}
Budzianowski, P., Vulic, I.: Hello, it’s gpt-2-how can i help you? towards
  the use of pretrained language models for task-oriented dialogue systems.
\newblock EMNLP-IJCNLP 2019 p.~15 (2019)

\bibitem{obstetrics-sonography}
Callen, P.W.: Ultrasonography in obstetrics and gynecology  (1988).
\newblock Publisher: WB Saunders CBS Educ. and Professional Publ., New York, NY

\bibitem{campbell_universal_2013}
Campbell, J., Dussault, G., Buchan, J., Pozo-Martin, F., Guerra, A.M., Leone,
  C.: A universal truth: no health without a workforce.
\newblock {WHO}, WHO, Recife, Brazil (2013)

\bibitem{text-based-behaviour-coding-forecasting}
Cao, J., Tanana, M., Imel, Z.E., Poitras, E., Atkins, D.C., Srikumar, V.:
  Observing dialogue in therapy: Categorizing and forecasting behavioral codes.
\newblock In: A.~Korhonen, D.R. Traum, L.~M{\`{a}}rquez (eds.) Proceedings of
  the 57th Conference of the Association for Computational Linguistics, {ACL}
  2019, Florence, Italy, July 28- August 2, 2019, Volume 1: Long Papers, pp.
  5599--5611. Association for Computational Linguistics (2019).
\newblock \doi{10.18653/v1/p19-1563}.
\newblock \urlprefix\url{https://doi.org/10.18653/v1/p19-1563}

\bibitem{proceeding5}
Chang, A., Dai, A., Funkhouser, T., Halber, M., Niessner, M., Savva, M., Song,
  S., Zeng, A., Zhang, Y.: Matterport3d: Learning from rgb-d data in indoor
  environments.
\newblock arXiv preprint arXiv:1709.06158  (2017)

\bibitem{ChaouaRCHH182}
Chaoua, I., Consoli, S., Harma, A., Helaoui, R., Recupero, D.R.: Analysis of
  topic propagation in therapy sessions using partially labeled latent
  dirichlet allocation.
\newblock In: Lecture Notes in Artificial Intelligence (2018)

\bibitem{DBLP:conf/ijcai/ChaouaRCHH18}
Chaoua, I., Recupero, D.R., Consoli, S., Harma, A., Helaoui, R.: Detecting and
  tracking ongoing topics in psychotherapeutic conversations.
\newblock In: Proceedings of the First Joint Workshop on {AI} in Health
  organized as part of the Federated {AI} Meeting {(FAIM} 2018), co-located
  with {AAMAS} 2018, {ICML} 2018, {IJCAI} 2018 and {ICCBR} 2018, Stockholm,
  Sweden, July 13-14, 2018., pp. 97--108 (2018).
\newblock \urlprefix\url{http://ceur-ws.org/Vol-2142/paper6.pdf}

\bibitem{constantin_end--end_2018}
Constantin, S., Niehues, J., Waibel, A.: An {End}-to-{End} {Goal}-{Oriented}
  {Dialog} {System} with a {Generative} {Natural} {Language} {Response}
  {Generation}.
\newblock In: {arXiv}:1803.02279 [cs]. Singapore (2018).
\newblock \urlprefix\url{http://arxiv.org/abs/1803.02279}.
\newblock ArXiv: 1803.02279

\bibitem{Crown2015Potential}
Crown, W.H.: Potential {Application} of {Machine} {Learning} in {Health}
  {Outcomes} {Research} and {Some} {Statistical} {Cautions}.
\newblock Value in Health \textbf{18}(2), 137--140 (2015).
\newblock Publisher: Elsevier BV

\bibitem{Damen2018EPIC}
Damen, D., Doughty, H., Farinella, G.M., Fidler, S., Furnari, A., Kazakos, E.,
  Moltisanti, D., Munro, J., Perrett, T., Price, W., Wray, M.: Scaling
  egocentric vision: The epic-kitchens dataset.
\newblock In: European Conference on Computer Vision (ECCV) (2018)

\bibitem{damen2022rescaling}
Damen, D., Doughty, H., Farinella, G.M., Furnari, A., Kazakos, E., Ma, J.,
  Moltisanti, D., Munro, J., Perrett, T., Price, W., et~al.: Rescaling
  egocentric vision: Collection, pipeline and challenges for epic-kitchens-100.
\newblock International Journal of Computer Vision pp. 1--23 (2022)

\bibitem{Damen2015}
Damen, D., Leelasawassuk, T., Mayol-Cuevas, W.: You-do, i-learn: Egocentric
  unsupervised discovery of objects and their modes of interaction towards
  video-based guidance.
\newblock Computer Vision and Image Understanding \textbf{149}, 98 -- 112
  (2016).
\newblock \doi{https://doi.org/10.1016/j.cviu.2016.02.016}.
\newblock
  \urlprefix\url{http://www.sciencedirect.com/science/article/pii/S1077314216000709}.
\newblock Special issue on Assistive Computer Vision and Robotics - Assistive
  Solutions for Mobility, Communication and HMI

\bibitem{David2020Evaluating}
{David Bellamy}, {Leo Celi}, {Andrew L. Beam}: Evaluating {Progress} on
  {Machine} {Learning} for {Longitudinal} {Electronic} {Healthcare} {Data}.
\newblock ArXiv  (2020)

\bibitem{dengImagenetLargescaleHierarchical2009}
Deng, J., Dong, W., Socher, R., Li, L.J., Li, K., Fei-Fei, L.: Imagenet: {A}
  large-scale hierarchical image database.
\newblock In: 2009 {IEEE} conference on computer vision and pattern
  recognition, pp. 248--255. Ieee (2009)

\bibitem{DBLP:conf/iui/DessiHKRR20}
Dess{\`{\i}}, D., Helaoui, R., Kumar, V., Recupero, D.R., Riboni, D.: {TF-IDF}
  vs word embeddings for morbidity identification in clinical notes: An initial
  study.
\newblock In: S.~Consoli, D.R. Recupero, D.~Riboni (eds.) Proceedings of the
  First Workshop on Smart Personal Health Interfaces co-located with 25th
  International Conference on Intelligent User Interfaces, SmartPhil@IUI 2020,
  Cagliari, Italy, March 17, 2020, \emph{{CEUR} Workshop Proceedings}, vol.
  2596, pp. 1--12. CEUR-WS.org (2020).
\newblock \urlprefix\url{http://ceur-ws.org/Vol-2596/paper1.pdf}

\bibitem{deviDesignImplementationAdvanced2022}
Devi, M.K., Vemuri, V.P., Arumugam, M., UmaMaheswaran, S.K., Acharjee, P.B.,
  Singh, R., Kaliyaperumal, K.: Design and {Implementation} of {Advanced}
  {Machine} {Learning} {Management} and {Its} {Impact} on {Better} {Healthcare}
  {Services}: {A} {Multiple} {Regression} {Analysis} {Approach} ({MRAA}).
\newblock Computational and Mathematical Methods in Medicine \textbf{2022},
  1--7 (2022).
\newblock Publisher: Hindawi Limited

\bibitem{devlin-etal-2019-bert}
Devlin, J., Chang, M.W., Lee, K., Toutanova, K.: {BERT}: Pre-training of deep
  bidirectional transformers for language understanding.
\newblock In: J.~Burstein, C.~Doran, T.~Solorio (eds.) Proceedings of the 2019
  Conference of the North {A}merican Chapter of the Association for
  Computational Linguistics: Human Language Technologies, Volume 1 (Long and
  Short Papers), pp. 4171--4186. Association for Computational Linguistics,
  Minneapolis, Minnesota (2019).
\newblock \doi{10.18653/v1/N19-1423}.
\newblock \urlprefix\url{https://aclanthology.org/N19-1423}

\bibitem{devlinBERTPretrainingDeep2019}
Devlin, J., Chang, M.W., Lee, K., Toutanova, K.: {BERT}: {Pre}-training of
  {Deep} {Bidirectional} {Transformers} for {Language} {Understanding} (2019).
\newblock \urlprefix\url{http://arxiv.org/abs/1810.04805}.
\newblock ArXiv:1810.04805 [cs]

\bibitem{feingold2018differences}
Feingold~Polak, R., Elishay, A., Shachar, Y., Stein, M., Edan, Y., Levy~Tzedek,
  S.: Differences between young and old users when interacting with a humanoid
  robot: A qualitative usability study.
\newblock In: Companion of the 2018 ACM/IEEE International Conference on
  Human-Robot Interaction, pp. 107--108. ACM (2018)

\bibitem{finkelstein_effectiveness_2016}
Finkelstein, E.A., Haaland, B.A., Bilger, M., Sahasranaman, A., Sloan, R.A.,
  Nang, E.E.K., Evenson, K.R.: Effectiveness of activity trackers with and
  without incentives to increase physical activity ({TRIPPA}): a randomised
  controlled trial.
\newblock The Lancet Diabetes \& Endocrinology \textbf{4}(12), 983--995 (2016).
\newblock \doi{10.1016/S2213-8587(16)30284-4}.
\newblock
  \urlprefix\url{/journals/landia/article/PIIS2213-8587(16)30284-4/abstract}

\bibitem{fitzpatrick_delivering_2017}
Fitzpatrick, K.K., Darcy, A., Vierhile, M.: Delivering {Cognitive} {Behavior}
  {Therapy} to {Young} {Adults} {With} {Symptoms} of {Depression} and {Anxiety}
  {Using} a {Fully} {Automated} {Conversational} {Agent} ({Woebot}): {A}
  {Randomized} {Controlled} {Trial}.
\newblock JMIR Mental Health \textbf{4}(2), e19 (2017).
\newblock \doi{10.2196/mental.7785}.
\newblock \urlprefix\url{https://mental.jmir.org/2017/2/e19/}

\bibitem{Furnari2018a}
Furnari, A., Battiato, S., Farinella, G.M.: Leveraging uncertainty to rethink
  loss functions and evaluation measures for egocentric action anticipation.
\newblock In: European Conference on Computer Vision Workshops on Egocentric
  Perception, Interaction and Computing (EPIC) (2018)

\bibitem{Furnari2018b}
Furnari, A., Battiato, S., Farinella, G.M.: Personal-location-based temporal
  segmentation of egocentric videos for lifelogging applications.
\newblock Journal of Visual Communication and Image Representation \textbf{52},
  1 -- 12 (2018).
\newblock \doi{https://doi.org/10.1016/j.jvcir.2018.01.019}.
\newblock
  \urlprefix\url{http://www.sciencedirect.com/science/article/pii/S1047320318300269}

\bibitem{Furnari2017c}
Furnari, A., Battiato, S., Grauman, K., Farinella, G.M.: Next-active-object
  prediction from egocentric videos.
\newblock Journal of Visual Communication and Image Representation \textbf{49},
  401 -- 411 (2017).
\newblock \doi{https://doi.org/10.1016/j.jvcir.2017.10.004}.
\newblock
  \urlprefix\url{http://www.sciencedirect.com/science/article/pii/S1047320317301967}

\bibitem{Furnari2017a}
Furnari, A., Farinella, G.M., Battiato, S.: Recognizing personal locations from
  egocentric videos.
\newblock IEEE Transactions on Human-Machine Systems \textbf{47}(1), 6--18
  (2017).
\newblock \doi{10.1109/THMS.2016.2612002}

\bibitem{g.kumarSurveyMachineLearning2016}
{G. Kumar}, {Rohit Kalra}: A survey on {Machine} {Learning} {Techniques} in
  {Health} {Care} {Industry}  (2016)

\bibitem{fred}
Gangemi, A., Presutti, V., Recupero, D.R., Nuzzolese, A.G., Draicchio, F.,
  Mongiov{\`{\i}}, M.: Semantic web machine reading with {FRED}.
\newblock Semantic Web \textbf{8}(6), 873--893 (2017).
\newblock \doi{10.3233/SW-160240}.
\newblock \urlprefix\url{https://doi.org/10.3233/SW-160240}

\bibitem{ganguliMachineLearningPursuit2020}
Ganguli, I., Gordon, W.J., Lupo, C., Sands-Lincoln, M., George, J., Jackson,
  G., Rhee, K., Bates, D.W.: Machine {Learning} and the {Pursuit} of
  {High}-{Value} {Health} {Care}.
\newblock NEJM Catalyst \textbf{1}(6) (2020).
\newblock Publisher: Massachusetts Medical Society

\bibitem{gehrmann_end--end_2018}
Gehrmann, S., Dai, F., Elder, H., Rush, A.: End-to-{End} {Content} and {Plan}
  {Selection} for {Natural} {Language} {Generation} (2018)

\bibitem{Gilbert2015market}
Gilbert, R., Goldstein, H., Hemingway, H.: The market in healthcare data.
\newblock BMJ (Clinical research ed.) p. h5897 (2015).
\newblock Publisher: BMJ

\bibitem{guSupervisedLearningPervasive2023}
Gu, X., Deligianni, F., Han, J., Liu, X., Chen, W., Yang, G.Z., Lo, B.: Beyond
  {Supervised} {Learning} for {Pervasive} {Healthcare}.
\newblock IEEE Reviews in Biomedical Engineering pp. 1--21 (2023).
\newblock Publisher: Institute of Electrical and Electronics Engineers (IEEE)

\bibitem{Gurrin2014}
Gurrin, C., Smeaton, A.F., Doherty, A.R.: Lifelogging: Personal big data.
\newblock Found. Trends Inf. Retr. \textbf{8}(1), 1--125 (2014).
\newblock \doi{10.1561/1500000033}.
\newblock \urlprefix\url{http://dx.doi.org/10.1561/1500000033}

\bibitem{harma_probabilistic_2016}
H\"arm\"a, A., Helaoui, R.: Probabilistic scoring of validated insights for
  personal health programs.
\newblock In: Proc. {IEEE} {Symp}. {Series} {Comp}. {Int}. 2016. Athens, Greece
  (2016)

\bibitem{harutyunyanMultitaskLearningBenchmarking2019}
Harutyunyan, H., Khachatrian, H., Kale, D.C., Ver~Steeg, G., Galstyan, A.:
  Multitask learning and benchmarking with clinical time series data.
\newblock Scientific Data \textbf{6}(1) (2019).
\newblock Publisher: Springer Science and Business Media LLC

\bibitem{hofmann_efficacy_2012}
Hofmann, S.G., Asnaani, A., Vonk, I.J., Sawyer, A.T., Fang, A.: The {Efficacy}
  of {Cognitive} {Behavioral} {Therapy}: {A} {Review} of {Meta}-analyses.
\newblock Cognitive therapy and research \textbf{36}(5), 427--440 (2012).
\newblock \doi{10.1007/s10608-012-9476-1}.
\newblock \urlprefix\url{https://www.ncbi.nlm.nih.gov/pmc/articles/PMC3584580/}

\bibitem{houlsby-adapters}
Houlsby, N., Giurgiu, A., Jastrzebski, S., Morrone, B., De~Laroussilhe, Q.,
  Gesmundo, A., Attariyan, M., Gelly, S.: Parameter-efficient transfer learning
  for nlp.
\newblock In: International Conference on Machine Learning, pp. 2790--2799.
  PMLR (2019)

\bibitem{hudsonUnderstaffing2015}
Hudson, C.K., Shen, W.: Understaffing.
\newblock Organizational Psychology Review \textbf{5}(3), 244--263 (2015).
\newblock Publisher: SAGE Publications

\bibitem{jain2021embeddings}
Jain, N., Kalo, J.C., Balke, W.T., Krestel, R.: Do embeddings actually capture
  knowledge graph semantics?
\newblock In: European Semantic Web Conference, pp. 143--159. Springer (2021)

\bibitem{johnsonMIMICIVFreelyAccessible2023}
Johnson, A.E.W., Bulgarelli, L., Shen, L., Gayles, A., Shammout, A., Horng, S.,
  Pollard, T.J., Hao, S., Moody, B., Gow, B., Lehman, L.w.H., Celi, L.A., Mark,
  R.G.: {MIMIC}-{IV}, a freely accessible electronic health record dataset.
\newblock Scientific Data \textbf{10}(1), 1 (2023).
\newblock \doi{10.1038/s41597-022-01899-x}.
\newblock \urlprefix\url{https://www.nature.com/articles/s41597-022-01899-x}.
\newblock Number: 1 Publisher: Nature Publishing Group

\bibitem{Kanade2012}
Kanade, T., Hebert, M.: First-person vision.
\newblock Proceedings of the IEEE \textbf{100}(8), 2442--2453 (2012).
\newblock \doi{10.1109/JPROC.2012.2200554}

\bibitem{Kathrin2022Benchmark}
{Kathrin Blagec}, {Jakob Kraiger}, {Wolfgang Frühwirt}, {Matthias Samwald}:
  Benchmark datasets driving artificial intelligence development fail to
  capture the needs of medical professionals.
\newblock ArXiv  (2022)

\bibitem{proceeding8}
Kolve, E., Mottaghi, R., Han, W., VanderBilt, E., Weihs, L., Herrasti, A.,
  Deitke, M., Ehsani, K., Gordon, D., Zhu, Y., et~al.: Ai2-thor: An interactive
  3d environment for visual ai.
\newblock arXiv preprint arXiv:1712.05474  (2017)

\bibitem{sdaih23}
Kumar., V., Balloccu., S., Wu., Z., Reiter., E., Helaoui., R., Recupero., D.,
  Riboni., D.: Data augmentation for reliability and fairness in counselling
  quality classification.
\newblock In: Proceedings of the 1st Workshop on Scarce Data in Artificial
  Intelligence for Healthcare - SDAIH,, pp. 23--28. INSTICC, SciTePress (2023).
\newblock \doi{10.5220/0011531400003523}

\bibitem{kumar2023data}
Kumar, V., Balloccu, S., Wu, Z., Reiter, E., Helaoui, R., Recupero, D.R.,
  Riboni, D.: Data augmentation for reliability and fairness in counselling
  quality classification  (2023)

\bibitem{10.1007/978-3-031-37249-0_10}
Kumar, V., Medda, G., Recupero, D.R., Riboni, D., Helaoui, R., Fenu, G.: How do
  you feel? information retrieval in psychotherapy and fair ranking
  assessment.
\newblock In: L.~Boratto, S.~Faralli, M.~Marras, G.~Stilo (eds.) Advances in
  Bias and Fairness in Information Retrieval, pp. 119--133. Springer Nature
  Switzerland, Cham (2023)

\bibitem{9286431}
Kumar, V., Recupero, D.R., Riboni, D., Helaoui, R.: Ensembling classical
  machine learning and deep learning approaches for morbidity identification
  from clinical notes.
\newblock IEEE Access \textbf{9}, 7107--7126 (2021).
\newblock \doi{10.1109/ACCESS.2020.3043221}

\bibitem{9866735}
Kumar, V., Reforgiato~Recupero, D., Helaoui, R., Riboni, D.: K-lm: Knowledge
  augmenting in language models within the scholarly domain.
\newblock IEEE Access \textbf{10}, 91,802--91,815 (2022).
\newblock \doi{10.1109/ACCESS.2022.3201542}

\bibitem{law2020characterizing}
Law, P.M., Endert, A., Stasko, J.: Characterizing automated data insights.
\newblock In: 2020 IEEE Visualization Conference (VIS), pp. 171--175. IEEE
  (2020)

\bibitem{Lee2015}
Lee, Y.J., Grauman, K.: Predicting important objects for egocentric video
  summarization.
\newblock International Journal of Computer Vision \textbf{114}(1), 38--55
  (2015).
\newblock \doi{10.1007/s11263-014-0794-5}.
\newblock \urlprefix\url{https://doi.org/10.1007/s11263-014-0794-5}

\bibitem{Leo2017}
Leo, M., Medioni, G., Trivedi, M., Kanade, T., Farinella, G.: Computer vision
  for assistive technologies.
\newblock Computer Vision and Image Understanding \textbf{154}, 1 -- 15 (2017).
\newblock \doi{https://doi.org/10.1016/j.cviu.2016.09.001}.
\newblock
  \urlprefix\url{http://www.sciencedirect.com/science/article/pii/S1077314216301357}

\bibitem{li2019fine}
Li, F., Jin, Y., Liu, W., Rawat, B.P.S., Cai, P., Yu, H.: Fine-tuning
  bidirectional encoder representations from transformers (bert)--based models
  on large-scale electronic health record notes: an empirical study.
\newblock JMIR medical informatics \textbf{7}(3), e14,830 (2019)

\bibitem{autonomous-ultrasound-review}
Li, K., Xu, Y., Meng, M.Q.H.: An overview of systems and techniques for
  autonomous robotic ultrasound acquisitions.
\newblock IEEE Transactions on Medical Robotics and Bionics \textbf{3}(2),
  510--524 (2021).
\newblock \doi{10.1109/TMRB.2021.3072190}

\bibitem{liDeepReinforcementLearning2017}
Li, Y.: Deep reinforcement learning: {An} overview.
\newblock arXiv preprint arXiv:1701.07274  (2017)

\bibitem{lin2015learning}
Lin, Y., Liu, Z., Sun, M., Liu, Y., Zhu, X.: Learning entity and relation
  embeddings for knowledge graph completion.
\newblock In: Twenty-ninth AAAI conference on artificial intelligence (2015)

\bibitem{roberta}
Liu, Y., Ott, M., Goyal, N., Du, J., Joshi, M., Chen, D., Levy, O., Lewis, M.,
  Zettlemoyer, L., Stoyanov, V.: Roberta: {A} robustly optimized {BERT}
  pretraining approach.
\newblock CoRR \textbf{abs/1907.11692} (2019).
\newblock \urlprefix\url{http://arxiv.org/abs/1907.11692}

\bibitem{liventsevReinforcementLearningMessage2021}
Liventsev, V.: Reinforcement learning as message passing (2021).
\newblock \urlprefix\url{https://vadim.me/publications/mpdp/}.
\newblock Section: publications

\bibitem{liventsevIntensiveCareOne2024}
Liventsev, V., Fritz, T.: Intensive {Care} as {One} {Big} {Sequence} {Modeling}
  {Problem} (2024).
\newblock \doi{10.48550/arXiv.2402.17501}.
\newblock \urlprefix\url{http://arxiv.org/abs/2402.17501}.
\newblock ArXiv:2402.17501 [cs]

\bibitem{liventsevEffectivePatientSimulators2021}
Liventsev, V., Härmä, A., Petković, M.: Towards {Effective} {Patient}
  {Simulators}.
\newblock Frontiers in Artificial Intelligence \textbf{4} (2021).
\newblock
  \urlprefix\url{https://www.frontiersin.org/articles/10.3389/frai.2021.798659}

\bibitem{madotto_mem2seq:_2018}
Madotto, A., Wu, C.S., Fung, P.: Mem2seq: {Effectively} {Incorporating}
  {Knowledge} {Bases} into {End}-to-{End} {Task}-{Oriented} {Dialog} {Systems}.
\newblock arXiv:1804.08217 [cs]  (2018).
\newblock \urlprefix\url{http://arxiv.org/abs/1804.08217}.
\newblock ArXiv: 1804.08217

\bibitem{maityMachineLearningImproved2017}
Maity, N.G., Das, S.: Machine learning for improved diagnosis and prognosis in
  healthcare.
\newblock In: 2017 {IEEE} {Aerospace} {Conference}. IEEE (2017)

\bibitem{mann2015people}
Mann, J.A., MacDonald, B.A., Kuo, I.H., Li, X., Broadbent, E.: People respond
  better to robots than computer tablets delivering healthcare instructions.
\newblock Computers in Human Behavior \textbf{43}, 112--117 (2015)

\bibitem{Mann97}
Mann, S.: Wearable computing: a first step toward personal imaging.
\newblock Computer \textbf{30}(2), 25--32 (1997).
\newblock \doi{10.1109/2.566147}

\bibitem{mcdermottReproducibilityMachineLearning2021}
McDermott, M.B.A., Wang, S., Marinsek, N., Ranganath, R., Foschini, L.,
  Ghassemi, M.: Reproducibility in machine learning for health research:
  {Still} a ways to go.
\newblock Science Translational Medicine \textbf{13}(586) (2021).
\newblock Publisher: American Association for the Advancement of Science (AAAS)

\bibitem{mercerMessageEditorinChief2008}
Mercer, K.: A {Message} from the {Editor}-in-{Chief}.
\newblock Healthcare Management Forum \textbf{21}(2), 4--4 (2008).
\newblock Publisher: SAGE Publications

\bibitem{ultraprocessed-food-survey-eu}
Mertens, E., Colizzi, C., Pe{\~n}alvo, J.L.: Ultra-processed food consumption
  in adults across europe.
\newblock European journal of nutrition \textbf{61}(3), 1521--1539 (2022)

\bibitem{mi}
Miller, W.R., Rollnick, S.: Motivational interviewing: Helping people change.
\newblock Guilford press (2012)

\bibitem{miller_toward_2009}
Miller, W.R., Rose, G.S.: Toward a {Theory} of {Motivational} {Interviewing}.
\newblock The American psychologist \textbf{64}(6), 527--537 (2009).
\newblock \doi{10.1037/a0016830}.
\newblock \urlprefix\url{http://www.ncbi.nlm.nih.gov/pmc/articles/PMC2759607/}

\bibitem{mitraMachineLearningHealthcare2021}
Mitra, D., Paul, A., Chatterjee, S.: Machine {Learning} in {Healthcare}.
\newblock In: {AI} {Innovation} in {Medical} {Imaging} {Diagnostics}, pp.
  37--60. IGI Global (2021)

\bibitem{munnUnderstaffingWardsCompromising2017}
Munn, F.: Understaffing on wards ‘compromising safe care’.
\newblock Nursing Standard \textbf{31}(36), 9--9 (2017).
\newblock Publisher: RCN Publishing Ltd.

\bibitem{nguyen2018novel}
Nguyen, T.D., Nguyen, D.Q., Phung, D., et~al.: A novel embedding model for
  knowledge base completion based on convolutional neural network.
\newblock In: Proceedings of the 2018 Conference of the North American Chapter
  of the Association for Computational Linguistics: Human Language
  Technologies, Volume 2 (Short Papers), pp. 327--333 (2018)

\bibitem{nhs-malnutrition}
NHS: Malnutrition.
\newblock \url{https://www.nhs.uk/conditions/malnutrition/} (2020).
\newblock Accessed on 10/11/2022

\bibitem{nickel2016holographic}
Nickel, M., Rosasco, L., Poggio, T.: Holographic embeddings of knowledge
  graphs.
\newblock In: Proceedings of the AAAI Conference on Artificial Intelligence,
  vol.~30 (2016)

\bibitem{Ortis2017}
Ortis, A., Farinella, G.M., D’Amico, V., Addesso, L., Torrisi, G., Battiato,
  S.: Organizing egocentric videos of daily living activities.
\newblock Pattern Recognition \textbf{72}, 207 -- 218 (2017).
\newblock \doi{https://doi.org/10.1016/j.patcog.2017.07.010}.
\newblock
  \urlprefix\url{http://www.sciencedirect.com/science/article/pii/S0031320317302819}

\bibitem{Pahwa2021Big}
Pahwa, K., Chauhan, S.: Big {Data} and {Machine} {Learning} in {Healthcare}:
  {Tools} \&amp; {Challenges}.
\newblock In: 2021 3rd {International} {Conference} on {Advances} in
  {Computing}, {Communication} {Control} and {Networking} ({ICAC3N}). IEEE
  (2021)

\bibitem{palMachineLearningHealthcare2023}
Pal, D., Ghosh, A., Majumdar, S., Pan, A., Ghosh, D.: Machine {Learning} in
  {Healthcare}: {A} {Review}.
\newblock methods \textbf{12}(1), 60--66 (2023)

\bibitem{pianykhImprovingHealthcareOperations2020}
Pianykh, O.S., Guitron, S., Parke, D., Zhang, C., Pandharipande, P., Brink, J.,
  Rosenthal, D.: Improving healthcare operations management with machine
  learning.
\newblock Nature Machine Intelligence \textbf{2}(5), 266--273 (2020).
\newblock Publisher: Springer Science and Business Media LLC

\bibitem{purushothamBenchmarkingDeepLearning2018}
Purushotham, S., Meng, C., Che, Z., Liu, Y.: Benchmarking deep learning models
  on large healthcare datasets.
\newblock Journal of Biomedical Informatics \textbf{83}, 112--134 (2018).
\newblock Publisher: Elsevier BV

\bibitem{puzikov2018e2e}
Puzikov, Y., Gurevych, I.: E2e nlg challenge: Neural models vs. templates.
\newblock INLG 2018 p. 463 (2018)

\bibitem{r.stanleyUnderstaffedOverwhelmed2010}
{R. Stanley}: Understaffed and overwhelmed  (2010)

\bibitem{radford2018improving}
Radford, A., Narasimhan, K., Salimans, T., Sutskever, I.: Improving language
  understanding by generative pre-training  (2018)

\bibitem{proceeding6}
Ramakrishnan, S.K., Gokaslan, A., Wijmans, E., Maksymets, O., Clegg, A.,
  Turner, J., Undersander, E., Galuba, W., Westbury, A., Chang, A.X., et~al.:
  Habitat-matterport 3d dataset (hm3d): 1000 large-scale 3d environments for
  embodied ai.
\newblock arXiv preprint arXiv:2109.08238  (2021)

\bibitem{reedGeneralistAgent2022}
Reed, S., Zolna, K., Parisotto, E., Colmenarejo, S.G., Novikov, A.,
  Barth-Maron, G., Gimenez, M., Sulsky, Y., Kay, J., Springenberg, J.T.,
  Eccles, T., Bruce, J., Razavi, A., Edwards, A., Heess, N., Chen, Y., Hadsell,
  R., Vinyals, O., Bordbar, M., de~Freitas, N.: A {Generalist} {Agent} (2022).
\newblock \urlprefix\url{http://arxiv.org/abs/2205.06175}.
\newblock ArXiv:2205.06175 [cs]

\bibitem{reiter-dale-2000}
Reiter, E., Dale, R.: Building Natural Language Generation Systems.
\newblock Studies in Natural Language Processing. Cambridge University Press
  (2000).
\newblock \doi{10.1017/CBO9780511519857}

\bibitem{reynaEarlyPredictionSepsis2020}
Reyna, M.A., Josef, C.S., Jeter, R., Shashikumar, S.P., Westover, M.B., Nemati,
  S., Clifford, G.D., Sharma, A.: Early prediction of sepsis from clinical
  data: the {PhysioNet}/{Computing} in {Cardiology} {Challenge} 2019.
\newblock Critical care medicine \textbf{48}(2), 210--217 (2020).
\newblock ISBN: 0090-3493 Publisher: LWW

\bibitem{RiboniBCJH16}
Riboni, D., Bettini, C., Civitarese, G., Janjua, Z.H., Helaoui, R.: Smartfaber:
  Recognizing fine-grained abnormal behaviors for early detection of mild
  cognitive impairment.
\newblock Artificial Intelligence in Medicine \textbf{67}, 57--74 (2016)

\bibitem{rodin2022untrimmed}
Rodin, I., Furnari, A., Mavroeidis, D., Farinella, G.M.: Untrimmed action
  anticipation.
\newblock In: International Conference on Image Analysis and Processing, pp.
  337--348. Springer (2022)

\bibitem{rodin2023egocentric}
Rodin, I., Furnari, A., Mavroeidis, D., Farinella, G.M.: Egocentric action
  anticipation for personal health.
\newblock In: ICASSP 2023-2023 IEEE International Conference on Acoustics,
  Speech and Signal Processing (ICASSP), pp. 1--5. IEEE (2023)

\bibitem{rogers2020primer}
Rogers, A., Kovaleva, O., Rumshisky, A.: A primer in bertology: What we know
  about how bert works.
\newblock Transactions of the Association for Computational Linguistics
  \textbf{8}, 842--866 (2020)

\bibitem{mi-in-healthcare}
Rollnick, S., Miller, W.R., Butler, C.: Motivational interviewing in health
  care: helping patients change behavior.
\newblock Guilford Press (2008)

\bibitem{rubak_motivational_2005}
Rubak, S., Sandbæk, A., Lauritzen, T., Christensen, B.: Motivational
  interviewing: a systematic review and meta-analysis.
\newblock The British Journal of General Practice \textbf{55}(513), 305--312
  (2005).
\newblock \urlprefix\url{http://www.ncbi.nlm.nih.gov/pmc/articles/PMC1463134/}

\bibitem{Ryoo2016}
Ryoo, M.S., Matthies, L.: First-person activity recognition: Feature, temporal
  structure, and prediction.
\newblock International Journal of Computer Vision \textbf{119}(3), 307--328
  (2016).
\newblock \doi{10.1007/s11263-015-0847-4}.
\newblock \urlprefix\url{https://doi.org/10.1007/s11263-015-0847-4}

\bibitem{S2017Benchmark}
{S. Purushotham}, {Chuizheng Meng}, {Zhengping Che}, {Yan Liu}: Benchmark of
  {Deep} {Learning} {Models} on {Large} {Healthcare} {MIMIC} {Datasets}.
\newblock arXiv.org  (2017)

\bibitem{sacha2014knowledge}
Sacha, D., Stoffel, A., Stoffel, F., Kwon, B.C., Ellis, G., Keim, D.A.:
  Knowledge generation model for visual analytics.
\newblock IEEE transactions on visualization and computer graphics
  \textbf{20}(12), 1604--1613 (2014)

\bibitem{isoug-guidelines}
Salomon, L.J., Alfirevic, Z., Berghella, V., Bilardo, C., Hernandez-Andrade,
  E., Johnsen, S., Kalache, K., Leung, K.Y., Malinger, G., Munoz, H., {others}:
  Practice guidelines for performance of the routine mid-trimester fetal
  ultrasound scan.
\newblock Ultrasound in Obstetrics \& Gynecology \textbf{37}(1), 116--126
  (2011).
\newblock Publisher: John Wiley \& Sons, Ltd. Chichester, UK

\bibitem{saraiya2005insight}
Saraiya, P., North, C., Duca, K.: An insight-based methodology for evaluating
  bioinformatics visualizations.
\newblock IEEE transactions on visualization and computer graphics
  \textbf{11}(4), 443--456 (2005)

\bibitem{proceeding4}
Savva, M., Kadian, A., Maksymets, O., Zhao, Y., Wijmans, E., Jain, B., Straub,
  J., Liu, J., Koltun, V., Malik, J., et~al.: Habitat: A platform for embodied
  ai research.
\newblock In: Proceedings of the IEEE/CVF international conference on computer
  vision, pp. 9339--9347 (2019)

\bibitem{schlichtkrull2018modeling}
Schlichtkrull, M., Kipf, T.N., Bloem, P., Van Den~Berg, R., Titov, I., Welling,
  M.: Modeling relational data with graph convolutional networks.
\newblock In: European semantic web conference, pp. 593--607. Springer (2018)

\bibitem{obesity-act-scot}
Scotland, O.A.: Obesity in scotland - prevalence, causes and impact factsheets.
\newblock
  \url{https://www.obesityactionscotland.org/media/locdychb/obesity-prevalence-causes--impact-202122-data-f.pdf}
  (2022).
\newblock Accessed on 9/12/2022

\bibitem{shang2019end}
Shang, C., Tang, Y., Huang, J., Bi, J., He, X., Zhou, B.: End-to-end
  structure-aware convolutional networks for knowledge base completion.
\newblock In: Proceedings of the AAAI Conference on Artificial Intelligence,
  vol.~33, pp. 3060--3067 (2019)

\bibitem{Singh2016}
Singh, S., Arora, C., Jawahar, C.V.: First person action recognition using deep
  learned descriptors.
\newblock In: 2016 IEEE Conference on Computer Vision and Pattern Recognition
  (CVPR), pp. 2620--2628 (2016).
\newblock \doi{10.1109/CVPR.2016.287}

\bibitem{sun2019rotate}
Sun, Z., Deng, Z.H., Nie, J.Y., Tang, J.: Rotate: Knowledge graph embedding by
  relational rotation in complex space.
\newblock arXiv preprint arXiv:1902.10197  (2019)

\bibitem{susaiyah2022aberdeen}
Susaiyah, A., H{\"a}rm{\"a}, A., Balloccu, S., Reiter, E., Petkovi{\'c}, M.:
  Smart selection of useful insights from wearables  (2023).
\newblock \urlprefix\url{https://ieeexplore.ieee.org/document/10193140/}

\bibitem{susaiyah2020towards}
Susaiyah, A., H{\"a}rm{\"a}, A., Reiter, E., Helaoui, R., Petkovi{\'c}, M.,
  et~al.: Towards a generalised framework for behaviour insight mining.
\newblock In: SmartPHIL: 1st Workshop on Smart Personal Health Interfaces. ACM
  (2020)

\bibitem{susaiyah2021neural}
Susaiyah, A., H{\"a}rm{\"a}, A., Reiter, E., Petkovi{\'c}, M.: Neural scoring
  of logical inferences from data using feedback.
\newblock International Journal of Interactive Multimedia \& Artificial
  Intelligence \textbf{6}(5) (2021)

\bibitem{ecai-paper}
Susaiyah, A., Härmä, A., Reiter, E., Petković, M.: Iterative neural scoring
  of validated insight candidates.
\newblock In: ECAI workshop on Intelligent Information Processing and Natural
  Language Generation. Santiago de Compostela, Spain (2020).
\newblock
  \urlprefix\url{https://intellang.github.io/papers/6-IntelLanG\_2020\_paper\_6.pdf}

\bibitem{thelancetHealthcareSystemStaffing2018}
{The Lancet}: Health-care system staffing: a universal shortfall.
\newblock The Lancet \textbf{392}(10161), 2238 (2018).
\newblock Publisher: Elsevier BV

\bibitem{thimInitialAssessmentTreatment2012}
Thim, T., Krarup, N.H.V., Grove, E.L., Rohde, C.V., Løfgren, B.: Initial
  assessment and treatment with the {Airway}, {Breathing}, {Circulation},
  {Disability}, {Exposure} ({ABCDE}) approach.
\newblock International journal of general medicine pp. 117--121 (2012).
\newblock Publisher: Taylor \& Francis

\bibitem{tortorellaHealthcareTrendsChallenges2020}
Tortorella, G.L., Fogliatto, F.S., Mac Cawley~Vergara, A., Vassolo, R.,
  Sawhney, R.: Healthcare 4.0: trends, challenges and research directions.
\newblock Production Planning \& Control \textbf{31}(15), 1245--1260 (2020).
\newblock \doi{10.1080/09537287.2019.1702226}.
\newblock \urlprefix\url{https://doi.org/10.1080/09537287.2019.1702226}.
\newblock Publisher: Taylor \& Francis \_eprint:
  https://doi.org/10.1080/09537287.2019.1702226

\bibitem{trouillon2016complex}
Trouillon, T., Welbl, J., Riedel, S., Gaussier, {\'E}., Bouchard, G.: Complex
  embeddings for simple link prediction.
\newblock In: International conference on machine learning, pp. 2071--2080.
  PMLR (2016)

\bibitem{wangGLUEMultitaskBenchmark2018}
Wang, A., Singh, A., Michael, J., Hill, F., Levy, O., Bowman, S.R.: {GLUE}: {A}
  multi-task benchmark and analysis platform for natural language
  understanding.
\newblock arXiv preprint arXiv:1804.07461  (2018)

\bibitem{wang2017knowledge}
Wang, Q., Mao, Z., Wang, B., Guo, L.: Knowledge graph embedding: A survey of
  approaches and applications.
\newblock IEEE Transactions on Knowledge and Data Engineering \textbf{29}(12),
  2724--2743 (2017)

\bibitem{wang2014knowledge}
Wang, Z., Zhang, J., Feng, J., Chen, Z.: Knowledge graph embedding by
  translating on hyperplanes.
\newblock In: Proceedings of the AAAI Conference on Artificial Intelligence,
  vol.~28 (2014)

\bibitem{wise_activity_2016}
Wise, J.: Activity trackers, even with cash incentives, do not improve health.
\newblock BMJ \textbf{355}, i5392 (2016).
\newblock \doi{10.1136/bmj.i5392}.
\newblock \urlprefix\url{https://www.bmj.com/content/355/bmj.i5392}

\bibitem{van_wissen_optimization_2016}
van Wissen, A., H\"arm\"a, A., Cuba~Gyllensten, I., Helaoui, R., Lowet, D.:
  Optimization of automated health programs by simulating user behaviors and
  program effects.
\newblock In: Proc. {Measuring} {Behavior} '2016. Dublin, Ireland (2016)

\bibitem{annomi-journal}
Wu, Z., Balloccu, S., Kumar, V., Helaoui, R., Reforgiato~Recupero, D., Riboni,
  D.: Creation, analysis and evaluation of annomi, a dataset of
  expert-annotated counselling dialogues.
\newblock Future Internet \textbf{15}(3) (2023).
\newblock \doi{10.3390/fi15030110}.
\newblock \urlprefix\url{https://www.mdpi.com/1999-5903/15/3/110}

\bibitem{annomi-icassp}
Wu, Z., Balloccu, S., Kumar, V., Helaoui, R., Reiter, E., Recupero, D.R.,
  Riboni, D.: Anno-mi: {A} dataset of expert-annotated counselling dialogues.
\newblock In: {IEEE} International Conference on Acoustics, Speech and Signal
  Processing, {ICASSP} 2022, Virtual and Singapore, 23-27 May 2022, pp.
  6177--6181. {IEEE} (2022).
\newblock \doi{10.1109/ICASSP43922.2022.9746035}.
\newblock \urlprefix\url{https://doi.org/10.1109/ICASSP43922.2022.9746035}.
\newblock (\copyright 2022 IEEE. Reprinted, with permission.)

\bibitem{mi-action-forecasting}
Wu, Z., Helaoui, R., Recupero, D.R., Riboni, D.: Towards automated counselling
  decision-making: Remarks on therapist action forecasting on the annomi
  dataset.
\newblock In: H.~Ko, J.H.L. Hansen (eds.) Interspeech 2022, 23rd Annual
  Conference of the International Speech Communication Association, Incheon,
  Korea, 18-22 September 2022, pp. 1906--1910. {ISCA} (2022).
\newblock \doi{10.21437/Interspeech.2022-506}.
\newblock \urlprefix\url{https://doi.org/10.21437/Interspeech.2022-506}.
\newblock DOI: 10.21437/Interspeech.2022-506

\bibitem{xhaferraRoleMachineLearning2022}
Xhaferra, E., Ismaili, F.: The {Role} of {Machine} {Learning} in the
  {Healthcare} {Sector}: {A} {Roadmap} to the {Potential} {Prospects}.
\newblock In: 2022 {International} {Congress} on {Human}-{Computer}
  {Interaction}, {Optimization} and {Robotic} {Applications} ({HORA}). IEEE
  (2022)

\bibitem{proceeding7}
Xia, F., Zamir, A.R., He, Z., Sax, A., Malik, J., Savarese, S.: Gibson env:
  Real-world perception for embodied agents.
\newblock In: Proceedings of the IEEE conference on computer vision and pattern
  recognition, pp. 9068--9079 (2018)

\bibitem{xie2016representation}
Xie, R., Liu, Z., Jia, J., Luan, H., Sun, M.: Representation learning of
  knowledge graphs with entity descriptions.
\newblock In: Proceedings of the AAAI Conference on Artificial Intelligence,
  vol.~30 (2016)

\bibitem{Yazhini2019State}
Yazhini, K., Loganathan, D.: A {State} of {Art} {Approaches} on {Deep}
  {Learning} {Models} in {Healthcare}: {An} {Application} {Perspective}.
\newblock In: 2019 3rd {International} {Conference} on {Trends} in
  {Electronics} and {Informatics} ({ICOEI}). IEEE (2019)

\end{thebibliography}
\end{document}